\pdfoutput=1
\documentclass[
]{ceurart}

\usepackage{indentfirst}
\usepackage{listings}
\usepackage{amsmath}
\usepackage{amssymb}
\usepackage{booktabs}
\usepackage{epsfig}
\usepackage{enumitem}
\usepackage{multirow}
\usepackage{setspace}
\usepackage{microtype}
\usepackage[american]{babel}
\usepackage{url}
\usepackage{bm}
\usepackage{color}
\newcommand{\eg}{\emph{e.g.}}

\newcommand{\ie}{\emph{i.e.}}

\usepackage{amsmath}

\usepackage{times}
\usepackage{epsfig}
\usepackage{graphicx}
\usepackage{amsmath}
\usepackage{amssymb}
\usepackage{color}
\usepackage[title]{appendix}

\usepackage{tabulary,multirow,subfig,xspace,xcolor}
\usepackage{algorithm}
\usepackage{etoolbox}
\makeatletter
\AfterEndEnvironment{algorithm}{\let\@algcomment\relax}
\AtEndEnvironment{algorithm}{\kern2pt\hrule\relax\vskip3pt\@algcomment}
\let\@algcomment\relax
\newcommand\algcomment[1]{\def\@algcomment{\footnotesize#1}}
\renewcommand\fs@ruled{\def\@fs@cfont{\bfseries}\let\@fs@capt\floatc@ruled
  \def\@fs@pre{\hrule height.8pt depth0pt \kern2pt}%
  \def\@fs@post{}%
  \def\@fs@mid{\kern2pt\hrule\kern2pt}%
  \let\@fs@iftopcapt\iftrue}
\makeatother

\definecolor{citecolor}{HTML}{0071bc}

\newlength\savewidth

\makeatletter\renewcommand\paragraph{\@startsection{paragraph}{4}{\z@}
  {.5em \@plus1ex \@minus.2ex}{-.5em}{\normalfont\normalsize\bfseries}}\makeatother
\begin{document}

\copyrightyear{2022}
\copyrightclause{Copyright for this paper by its authors.
  Use permitted under Creative Commons License Attribution 4.0
  International (CC BY 4.0).}


\conference{CLEF 2022: Conference and Labs of the Evaluation Forum, 
    September 5--8, 2022, Bologna, Italy}


\title{ Explored An Effective Methodology for Fine-Grained Snake Recognition}

\author[1]{Yong Huang}[%
email=yonghuang.hust@gmail.com,
]

\address{Huazhong University of Science and Technology,
   Wuhan, 430074, China}
\address{Alibaba Group, Ant Group, MyBank, Hangzhou, 310013, China}

\author[2]{Aderon Huang}[%
email=xugan.hy@alibaba-inc.com,
url=https://github.com/AderonHuang,
]

\author[3]{Wei Zhu}[%
email=wei.wz@antgroup.com,
]

\author[4]{Yanming Fang}[%
email=yanming.fym@antgroup.com,
]

\author[5]{Jinghua Feng}[%
email=jinghua.fengjh@alibaba-inc.com,
]

\begin{abstract}
Fine-Grained Visual Classification (FGVC) is a longstanding and fundamental problem in computer vision and pattern recognition, and underpins a diverse set of real-world applications. This paper describes our contribution at SnakeCLEF2022 with FGVC. Firstly, we design a strong multimodal backbone to utilize various meta-information to assist in fine-grained identification. Secondly, we provide new loss functions to solve the long tail distribution with dataset. Then, in order to take full advantage of unlabeled datasets, we use self-supervised learning and supervised learning joint training to provide pre-trained model. Moreover, some effective data process tricks also are considered in our experiments. Last but not least, fine-tuned in downstream task with hard mining, ensambled kinds of model performance. Extensive experiments demonstrate that our method can effectively improve the performance of fine-grained recognition. Our method can achieve a macro $f1$ score 92.7$\%$ and 89.4$\%$ on private and public dataset, respectively, which is the 1st place among the participators on private leaderboard. The code will be made available at \url{https://github.com/AderonHuang/fgvc9_snakeclef2022}.
\end{abstract}

\begin{keywords}
  Fine-Grained Visual Classification (FGVC) \sep
  Multimodal Backbone \sep
  Long Tail Distribution \sep
  Self-Supervised Learning (SSL) \sep
  Hard-Mining
\end{keywords} 

\maketitle

\section{Introduction}

The human visual system is naturally capable of fine-grained image reasoning, likely it not only distinguishes a dog from a wolf, but also know the difference between a Sea Snake and a Land Snake
. Fine-grained visual classification (FGVC) was introduced to the
academic community for the very same purpose, to teach how to ``\verb|see|'' with machine in a fine-grained manner. FGVC approaches are present in a wide-range of applications in both industry and research, with examples including automatic biodiversity monitoring, improving  eco-epidemiological data ~\cite{nabirds15,inat2017,inat2021}, and have resulted in a positive impact in area such as conservation ~\cite{FGICCVworkshop} and commerce ~\cite{wei2020deep}.

\begin{figure}[t!]
\centering
{\includegraphics[width=0.95\columnwidth]{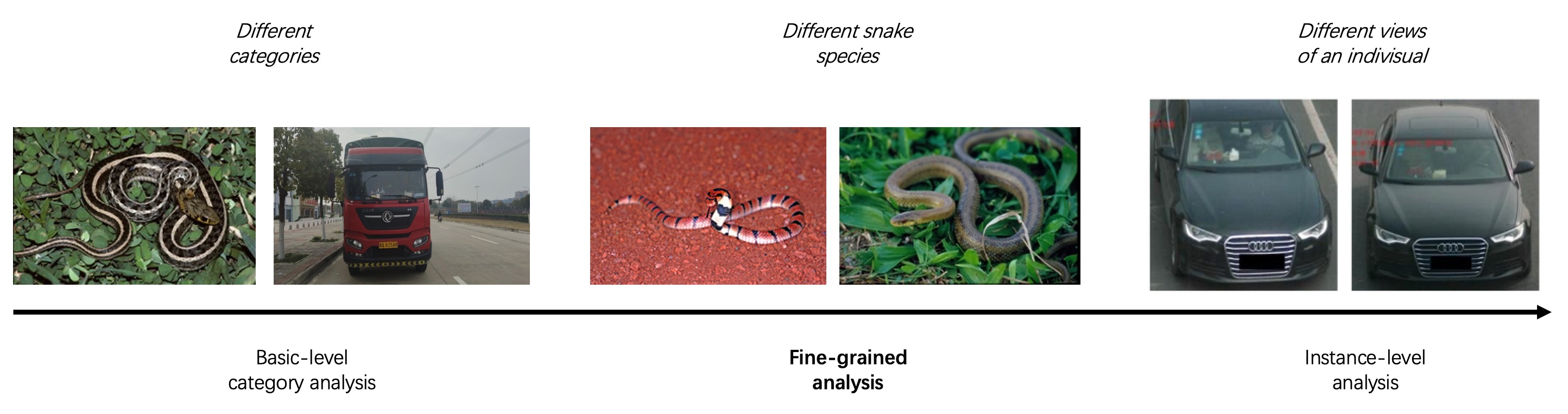}}
\vspace{-1em}
\caption{An illustration of fine-grained image recognition which lies in the continuum between the basic-level category analysis (\ie, generic image analysis) and the instance-level analysis.}
\label{fig:multigrained}
\end{figure}

\begin{figure}[b]
\centering
{\includegraphics[width=1.0\columnwidth]{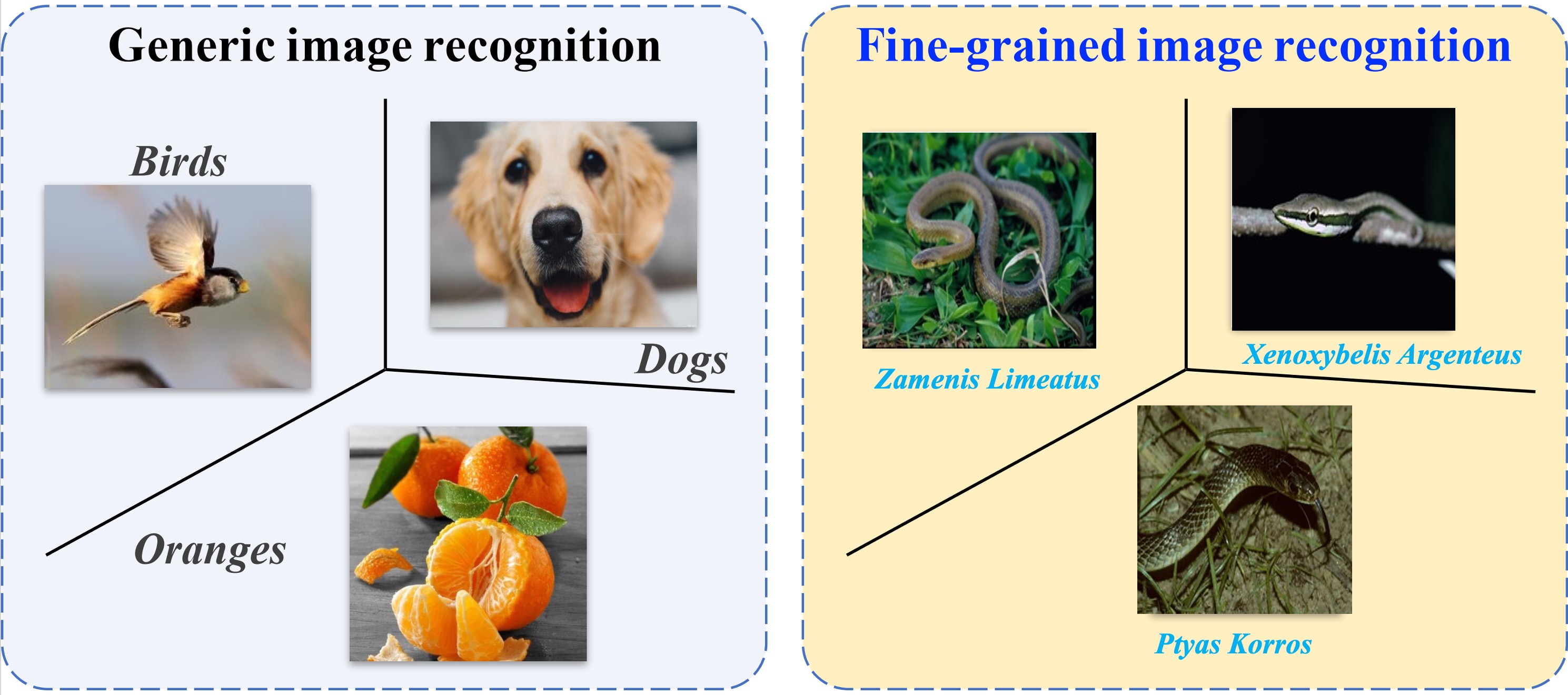}}
\vspace{-1em}
\caption{Generic image analysis vs fine-grained image recognition (using
visual classification as an example).}
\label{fig:3}
\end{figure}

Fine-grained visual classification (FGVC) focuses on dealing with objects belonging to multiple subordinate categories of the same meta-category (e.g., different species of snakes or
different models of sorghums). As illustrated in Figure~\ref{fig:multigrained}, fine-grained analysis lies in the continuum between basic-level category analysis (i.e., generic image analysis) and instance-level analysis (e.g., the identification of individuals). Particularly, what differes FGVC from generic image analysis is that target objects be classified to coarse-grain meta-categories in generic image analysis and thus are significantly different (e.g., determining if an image contains a bird, a fruit, or a dog). However, in FGVC, since objects typically come from subcategories of the same meta-category, the fine-grained nature of the problem causes them to be visually similar~\cite{wei2021fine}. As an example of FGVC task, in Figure~\ref{fig:3}, the task
is to classify different species of snakes. For accurate image recognition, it is necessary to capture the subtle visual differences (e.g., discriminative features such as length, body shape or body color). Furthermore, as noted earlier, the fine-grained nature of the problem is challenging because of the small inter-class variations caused by highly similar sub-categories, and the large intra-class variations in background, scale, rotations. It is such as the opposite of generic image analysis (i.e., the small intra-class variations
and the large inter-class variations), and what makes FGVC a unique and challenging problem. 

\textbf{Formule:}In generic image recognition, we are given a training dataset $\mathcal{D}=\left\{ \left( \bm{x}^{(n)}, y^{(n)}\right) | i=1, ..., N  \right\}$, containing multiple images and associated class labels (\ie, $\bm{x}$ and $y$), where $y\in[1, ..., C]$. 
Each instance $\left(\bm{x}, y\right)$ belongs to the joint space of both the image and label spaces (\ie, $\mathcal{X}$ and $\mathcal{Y}$, respectively), according to the distribution of $p_r(\bm{x}, y)$
\begin{equation}
\left(\bm{x}, y\right) \in \mathcal{X}\times \mathcal{Y}\,.
\end{equation}
In particular, the label space $\mathcal{Y}$ is the union space of all the $C$ subspaces corresponding to the $C$ categories, \ie, $\mathcal{Y} = \mathcal{Y}_1 \cup \mathcal{Y}_2 \cup \cdots \cup \mathcal{Y}_c \cup \cdots \cup \mathcal{Y}_C$. Then, we can train a predictive/recognition deep network $f(\bm{x};\theta)$ parameterized by $\theta$ for generic image recognition by minimizing the expected risk
\begin{equation}
\min_\theta \mathbb{E}_{(\bm{x},y)\sim p_r(\bm{x}, y)} \left[ \mathcal{L}(y, f(\bm{x};\theta))\right]\,,
\end{equation}
where $\mathcal{L}(\cdot,\cdot)$ is a loss function that measures the match between the true labels and those predicted by $f(\cdot;\theta)$. 
While, as aforementioned, fine-grained recognition aims to accurately classify instances of different subordinate categories from a certain meta-category, \ie,
\begin{equation}
\left(\bm{x}, y'\right) \in \mathcal{X}\times \mathcal{Y}_c\,,
\end{equation}
where $y'$ denotes the fine-grained label and $\mathcal{Y}_c$ represents the label space of class $c$ as the meta-category. Therefore, the optimization objective of fine-grained recognition is as
\begin{equation}
\min_\theta \mathbb{E}_{(\bm{x},y')\sim p'_r(\bm{x}, y')} \left[ \mathcal{L}(y', f(\bm{x};\theta))\right]\,.
\end{equation} 

\section{Related Work}

The existing Fine-Grained Visual Classification (FGVC) methods can be divided into image only and multi-modality. The former relies entirely on visual information to tackle the problem of fine-grained classification, while the latter tries to take multi-modality data to establish joint representations for incorporating multi-modality information, facilitating finegrained
recognition.

\textbf{Image Only:}Fine-Grained Visual Classification (FGVC) methods that only rely on image can be roughly classified into two categories: localization methods~\cite{16,25,55} and featureencoding methods~\cite{52,54,57}. Early work ~\cite{27,48} used part annotations as supervision to make the network pay attention to the subtle discrepancy between some species and suffers from its expensive annotations. RA-CNN~\cite{15} was proposed to zoom in subtle regions, which recursively learns discriminative region attention and region-based feature representation at multiple scales in a mutually reinforced way. MA-CNN~\cite{53} designed a multi-attention module where part generation and feature learning can reinforce each other. NTSNet~\cite{51} proposed a self-supervision mechanism to localize informative regions without part annotations effectively. Feature-encoding methods are devoted to enriching feature expression capabilities to improve the performance of fine-grained classification. Bilinear CNN~\cite{24} was proposed to extract higher-order features, where two feature maps are multiplied using the outer product. HBP ~\cite{52} further designed a hierarchical framework to do crosslayer bilinear pooling. DBTNet~\cite{54} proposed deep bilinear transformation, which takes advantage of semantic information and can obtain bilinear features efficiently. CAP~\cite{2} designed context-aware attentional pooling to captures subtle changes in image. TransFG~\cite{18} proposed a Part Selection Module to select discriminative image patches applying vision transformer. Compared with localization methods, feature-encoding methods are difficult to tell us the discriminative regions between different species explicitly.

\textbf{Multimodality:}In order to differentiate between these challenging visual categories, it is helpful to take advantage of additional information, i.e., geolocation, attributes, and text description. Geo-Aware~\cite{6} introduced geographic information prior to fine-grained classification and systematically examined a variety of methods using geographic information prior, including post-processing, whitelisting, and feature modulation. Presence-Only~\cite{28} also introduced spatio-temporal prior into the network, proving that it can effectively improve the final classification performance. KERL~\cite{4} combined rich additional information and deep neural network architecture, which organized rich visual concepts in the form of a knowledge graph. Meanwhile, KERL~\cite{4} used a gated graph neural network to propagate node messages through the graph to generate knowledge representation. CVL~\cite{20} proposed a two-branch network where one branch learns visual features, one branch learns text features, and finally combines the two parts to obtain the final latent semantic representations. The methods mentioned above are all designed for specific prior information and cannot flexibly adapt to different auxiliary information.

\section{Method}

Analyze dataset firstly, we introduce the strong multimodal framework, and propose a new loss funcition to solve long tail distribution with dataset. Then, to make use of enormous unlabeled data, we design a self-supervised leaning framework. It can joint train with supervised learning framework by using joint loss function, as well as completely self-supervise leaning to provide pre-trained model. Furthermore, we also consider hard mining methods to improve the robustness and generalization of model. Last but not least, some tricks are used to improve performance. Eventually, models are ensambled by different ways to achieve maximum performance. 

\subsection{Dataset}

\begin{figure}[b]
\centering
{\includegraphics[width=1.0\columnwidth]{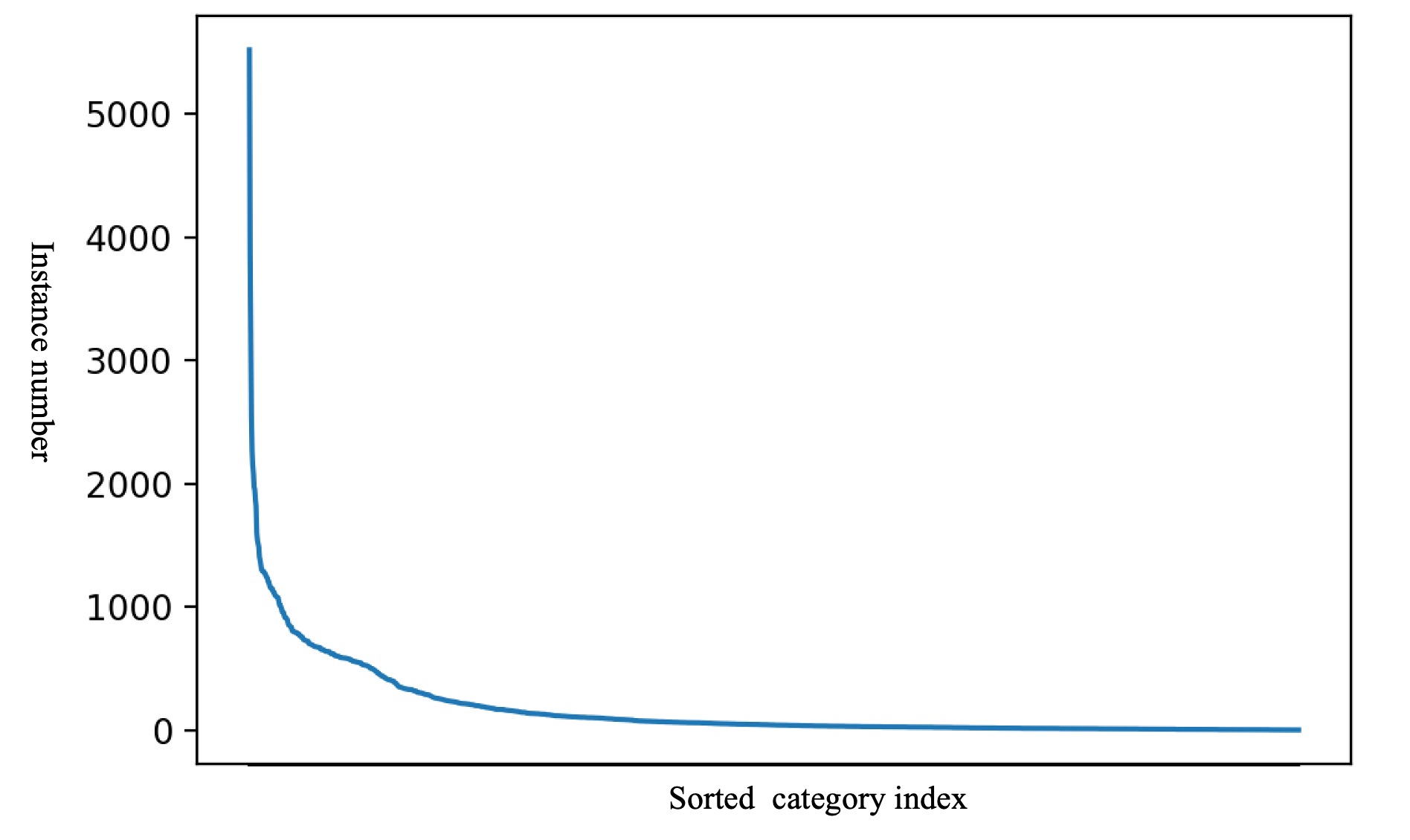}}
\vspace{-1em}
\caption{Insight for the number of each class in the training set.}
\label{fig:44}
\end{figure}

Based on the metadata file presented by the organizers, we concluded that the data at hand is highly imbalanced, which can be considered as the question of long tail distribution.  Since this could have negatively influenced our models, we need use weighted loss to counteract this phenomenon. Figure~\ref{fig:44} provides an insight for the number to each class in the training set. 

\subsection{Multimodal Framework}

\begin{figure}[b]
\centering
{\includegraphics[width=1.0\columnwidth]{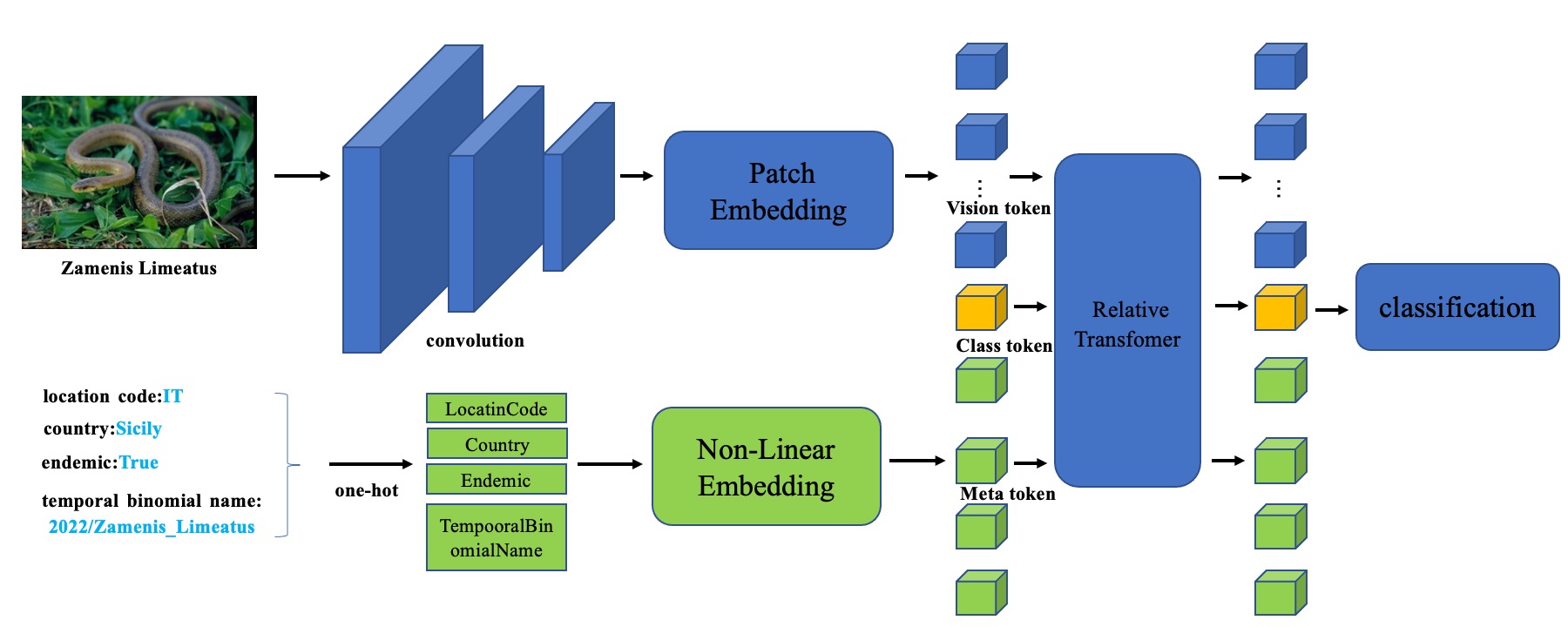}}
\vspace{-1em}
\caption{The overall multimodal framework, including visual information and text information.}
\label{fig:overview}
\end{figure}

We took MetaFormer~\cite{diao2022metaformer} and ConvNext~\cite{liu2022convnet} as our initial baseline. As shown in Figure~\ref{fig:overview}, MetaFormer is a mixed and multimodal framework that combines ConvNet and Transformer. It supports a simple and effectivie way for adding meta-information using transformer block. In our approach, we use MetaFormer and modify the input of meta-information, specially, images initialy input convolution to encode vision token, and use one-hot encoding to encode attribute meta-information such as location code, country, endemic and temporal binomial name to meta token. Vision token, Class token, Meta token are used for information fusion through the Relative Transformer layer. We also consider ConvNext as another network architecture to enhance the model diversity for later model ensemble, which is a pure ConvNet and strong feature extraction network. We apply hyperparameter fine-tuning to improve their performance. Some experimental details would illustrate the later ablation studies in Sec 4. 

\subsection{Effective Long-Tailed Loss}

The classifier trained by the widely applied Cross-Entropy (CE) Loss~\cite{ce} is highly biased on longtailed datasets, resulting in much lower accuracy of tail classes than head classes. The major reason is that gradients brought by positive samples are overwhelmed by gradients from negative samples in tail classes. We know Arcface Loss~\cite{arcface}, Seesaw Loss~\cite{seesaw}, Polyloss~\cite{polyloss}, Circle Loss~\cite{circle}, Additive Margin Softmax Loss~\cite{additive} that are applied to solve long-tailed distribution. We also attempt hard mining method  to improve model performance. We experiment these loss functions, later in this section, and detailed these works.

\textbf{Arcface Loss:}The most widely used classification loss function, softmax loss, is presented as following:
\begin{equation}
L_{softmax}(x)=-\frac{1}{N}\sum_{i=1}^{N}\log\frac{e^{W^T_{y_i} x_i+b_{y_i}}}{\sum_{j=1}^{n}e^{W^T_j x_i+b_j}},
\label{eq:softmax}
\vspace{-1mm}
\end{equation}
where $x_i\in\mathbb{R}^d$ denotes the deep feature of the $i$-th sample, belonging to the $y_i$-th class. The embedding feature dimension $d$ is set to $512$ in this paper. $W_j\in\mathbb{R}^d$ denotes the $j$-th column of the weight $W \in \mathbb{R}^{d \times n}$ and $b_j\in\mathbb{R}^n $ is the bias term. The batch size and the class number are $N$ and $n$, respectively. 
Arcface loss is presented as following~\cite{arcface}:
\begin{equation}
{L_{arcface}(\theta_{})}=-\frac{1}{N}\sum_{i=1}^{N}\log\frac{e^{s(\cos(\theta_{y_i}+m))}}{e^{s(\cos(\theta_{y_i}+m))}+\sum_{j=1,j\neq  y_i}^{n}e^{s\cos\theta_{j}}}+l_{2}.
\label{eq:arcface}
\vspace{-1mm}
\end{equation}
where $s$ is a radius of distributed on a hypersphere for learned embedding features, $\theta_j$ is the angle between the weight $W_j$ and the feature $x_i$, $m$ is an additive angular margin penalty between $x_i$ and $W_{y_i}$ to simultaneously enhance the intra-class compactness and inter-class discrepancy, $l_{2}$ is also known as the Euclidean norm to avoid overfitting. Our experiments show that arcface loss perform well.  

\textbf{Seesaw Loss:}dynamically re-balances positive and negative gradients for each catogory with two complementary factors, i.e., mitigation factor and compensation factor. Seesaw Loss rewrite the Cross Entropy loss as~\cite{seesaw}:
\begin{equation} \label{eql:seesawloss}
	\begin{aligned}
		L_{seesaw}(z)=-\sum_{i=1}^{N} y_{i} \log (\widehat{\sigma}_{i})+l_{2},\\
		\text{  with  } \widehat{\sigma}_{i}=\frac{e^{z_{i}}}{\sum_{j\neq i}^{N}S_{{ij}}e^{z_{j}}+e^{z_{i}}}.
	\end{aligned}
\end{equation}
where $z=[z_1, z_2, \dots, z_N]$ are predicted logits and $sigma = [\sigma_1, \sigma_2, \dots, \sigma_N]$ are probabilities of the classifier. And $y_i \in \{0,1\},	1\leq i \leq N$ is the one-hot ground truth label. $l_{2}$ is also known as the Euclidean norm to avoid overfitting. Seesaw loss determines $S_{ij}$ by a mitigation factor and a compensation factor, $\gamma$ is temperature coefficient, as we can see
\begin{equation}
	S_{i j} = \gamma*M_{i j} \cdot C_{ij}.
\end{equation}

\noindent
i) Mitigation Factor. Seesaw Loss accumulates instance number $D_i$ for each category $i$ at each iteration in the whole training process. given an instance with positive label $i$, for another category $j$,
the mitigation factor adjusts the penalty for negative label $j$ the ratio $\frac{D_j}{D_i}$
\begin{equation} \label{eql:class-wise}
	\begin{aligned}
		M_{i j} =\left\{\begin{array}{ll}
			\quad 1,                        & \text { if } D_i \leq D_j \\
			\left(\frac{D_j}{D_i}\right)^p, & \text { if } D_i > D_j
		\end{array}\right.
	\end{aligned}
\end{equation}
ii) Compensation factor. this factor compensates the diminished gradient when there is misclassification, \ie, the predicted probability $\sigma_j$ of negative label $j$ is greater than $\sigma_i$.
The compensation factor $C_{ij}$ is calculated as
\begin{equation} \label{eql:ins-wise}
	\begin{aligned}
		C_{i j} =\left\{\begin{array}{ll}
			\quad 1,                                  & \text { if } \sigma_j \leq \sigma_i \\
			\left(\frac{\sigma_j}{\sigma_i}\right)^q, & \text { if } \sigma_j > \sigma_i
		\end{array}\right.
	\end{aligned}
\end{equation}
the exponents $p$ and $q$ is hyper-parameters that control the scale, in our experiments, we empirically set $p$ = 0.8, $q$ = 2, $\gamma$ = 0.95, respectively.

\subsection{Self-Supervised Learning}
We consider unlabel of the test dataset, to fully mine unlabel data information. We use MoCo Method-based and Simclr Method-based to learn more robust and generalizable performance to compensate for the lack of supervised learning.  

\textbf{MoCo Method-based}:Pre-trained model through self-supervision learning, and fine-tune model use of labeled data. As common practice (\eg, \cite{He2020,Chen2020}), we take two crops for each image under random data augmentation. They are encoded by two encoders, $f_q$ and $f_k$, with output vectors $q$ and $k$. Intuitively, $q$ behaves like a ``query'' \cite{He2020}, and the goal of learning is to retrieve the corresponding ``key''. This is formulated as minimizing a contrastive loss function \cite{Hadsell2006}. We adopt the form of InfoNCE \cite{Oord2018}:
\begin{equation}
\small
\mathcal{L}_{q} = -\log \frac{\exp(q{\cdot}k^+ / \tau)}{\exp(q{\cdot}k^+ / \tau) + {\displaystyle\sum_{k^-}}\exp(q{\cdot}k^-  / \tau)}.
\label{eq:infonce}
\end{equation}
Here $k_{+}$ is $f_k$'s output on the same image as $q$, known as $q$'s positive sample. The set $\{k^{-}\}$ consists of $f_k$'s outputs from other images, known as $q$'s negative samples.
$\tau$ is a temperature hyper-parameter \cite{2018} for $\ell_2$-normalized $q$, $k$. We empirically set 0.25 in later experiments. Algorithm 1 show its pseudocode.

\begin{algorithm}[t]
\caption{MoCo Method-based: Pseudocode of in a python-like style. }
\label{alg:code}
\definecolor{codeblue}{rgb}{0.25,0.5,0.5}
\definecolor{codekw}{rgb}{0.85, 0.18, 0.50}
\begin{lstlisting}[language=python]
# f_q: encoder representation feature: backbone + proj mlp + pred mlp
# f_k: momentum encoder representation feature: backbone + proj mlp
# m: momentum coefficient
# tau: temperature

for x in loader:  # load a minibatch x with N samples
    x1, x2 = aug(x), aug(x)  # augmentation
    q1, q2 = f_q(x1), f_q(x2)  # queries: [N, C] each
    k1, k2 = f_k(x1), f_k(x2)  # keys: [N, C] each

    loss = ctr(q1, k2) + ctr(q2, k1)  # symmetrized
    loss.backward()
    
    update(f_q)  # optimizer update: f_q
    f_k = m*f_k + (1-m)*f_q  # momentum update: f_k
  
# contrastive loss
def ctr(q, k):
    logits = mm(q, k.t())  # [N, N] pairs
    labels = range(N)  # positives are in diagonal
    loss = CrossEntropyLoss(logits/tau, labels) 
    return 2 * tau * loss
\end{lstlisting}
\end{algorithm}
\textbf{Simclr Method-based}:It likes to weakly supervised learning, we use unlabeled data and labeled data joint training to provide pre-trained model. Then, we alse use labeled data to fine-tune classfied head. Its pre-trianed loss can be written as:
\begin{equation} \label{eql:seesawloss}
	\begin{aligned}
		L_{joint-loss}=\lambda_{1}*L_{sup}+\lambda_{2}*L_{self-sup}.
	\end{aligned}
\end{equation}
$\lambda_{1}$ and $\lambda_{2}$ are the supervised and self-supervised learning balance coefficients, respectively, which mainly adjust the balance stability in supervised and self-supervised training, and avoid the phenomenon of separately dominant learning. $L_{sup}$ and $L_{self-sup}$ are supervised loss and self-supervised loss, respectively. We emperically set 0.9 and 0.1 through multiple experiments respectively. Algorithm 2 shows its pseudocode. 
\begin{algorithm}[t]
\caption{Simclr Method-based: Pseudocode of in a python-like style. }
\label{alg:code}
\definecolor{codeblue}{rgb}{0.25,0.5,0.5}
\definecolor{codekw}{rgb}{0.85, 0.18, 0.50}
\begin{lstlisting}[language=python]
# f is network representation, like Metaformer-2,  

for x in loader1:  # load a unlabeled minibatch x with N samples
    for x_k in x:
        # draw two different augmentation functions, consider as aug1 and aug2
        # the first augmentation
        x_(2k-1) = aug1(x_k)
        h_(2k-1) = f(x_(2k-1))                                  #representation
        z_(2k-1) = g(h_(2k-1))                                      #projection
        
        # the second augmentation
        x_(2k) = aug1(x_k)
        h_(2k) = f(x_(2k))                                      #representation
        z_(2k) = g(h_(2k))                                          #projection
    for i,j in zip(z_(2k-1),z_(2k)):
        s(i,j) = z(i)*z(j)/(||z(i)|| ||z(j)||)
    
    # update networks f and g to minimize loss
    loss_(self-sup) = InfoNCE(s(i,j)) 
      
for xx in loader2: #load a labeled minibatch x with N samples
    xx1 = aug(xx)
    h = f(xx1)                                                 #representation
    z = g(h)                                                              #mlp
    loss_(sup) = CrossEntropy(z)
def ctr(loss_self-sup, loss_sup):
    loss = p * loss_sup + q * loss_self-sup
    return loss
\end{lstlisting}
\end{algorithm}

\subsection{Pre-Post-Process}
\begin{figure}[t!]
\centering
{\includegraphics[width=0.95\columnwidth]{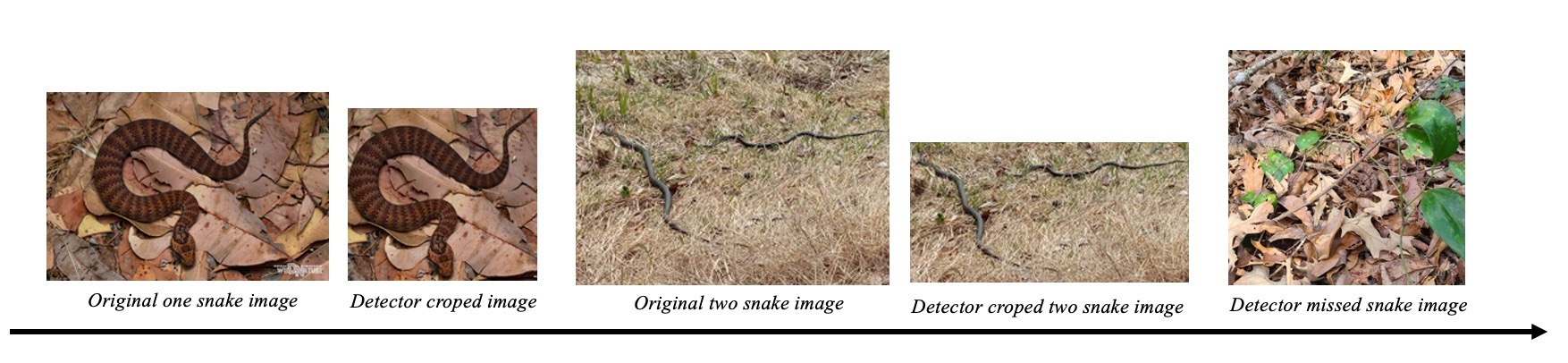}}
\vspace{-1em}
\caption{oringinal image, detector croped image and detector missed image.}
\label{fig:6}
\end{figure}

\textbf{Pre-Process. }Through statistical analysis of the training data, we observed that the foreground is small in the background, far from the center of the image, or there is not foreground. Therefore, in order to ensure that the foreground handles the image center position as much as possible, we use the target detector to crop out the foreground objects in the training set and test set. For images without foreground or images that are missed by the detector, the original scale is retained, and then send to the classifier. We show in Figure~\ref{fig:6}, oringinal image and croped image.

\textbf{Post-Process. }We have use center crop, five crop and multi scale ten crop during test phase, the effects of test time augmentation (TTA). It should be noted that the mean $f1$ score on public test set is out of accord with private test set in some experiments, and it is hard to decide which TTA is better only based on public score, in consideration of robustness, we have chosen multi scale five crop based on the public mean $f1$. Some experiments are shown later section. We finally use differenct models with their logits. It is acknowledged that the tail categories tend to have lower logits compared to head categories, so the tail categories is easier to be misclassified as head categories. As shown in Table \ref{table:7}, with our ensambling for tail categories, the mean $f1$ score improves a lot on private test set.

\section{Experiments}

In this section, we first elaborate on the implementation and training details. Then we introduce
ablation studies on loss functions, self-supervised learning and  bag of training settings, which improve our single model’s performance gradually. Then we list some other attempts and their results.

\subsection{Implementation Details}
We trained the model on SnakeCLEF2022 dataset~\cite{snakeclef2022} which contains of 270,251 training images
belonging to 1,572 species observed from global wide. The dataset has been divided into train
and val set, we use both train and val for training in most settings. We report the results on test set which totally contains 48,280 observations with corresponding images, after removing duplication of observations is 28,431. The test set
is divided into two parts. The public set contains 20\% of the data, and the private set contains 80\% of
the data. We conduct all the experiments with Tesla A100 (80G) and Tesla V100 (32G). We use AdamW optimizer with cosine learning scheduler, initialize the learning rate to $5e^{-5}$ and scale it by batch size, we include most of the augmentation and regularization strategies of~\cite{liu2021swin} in training, augmentation strategies include random erasing with probability, regularization strategies like to set empirical coefficients for supervised loss and self-supervised loss. 

\subsection{Ablation Studies}
As show in Table \ref{table:1}, We trained MetaFormer-2 for 300 epochs, with Soft Target Cross Entropy loss and mixup augmentation to build our baseline. For ablation studies, it should be noted that except the parameter to be compared, there are little other not consistent parameters. such as the accumulate steps in last row in Table \ref{table:2}, we argue that it will not affects the conclusion largely, therefore, we set the default accumulate steps is 2 in our later experiments.

\begin{table}[htbp]
\caption{MetaFormer-2 baseline.}
\label{table:1}
\setlength{\tabcolsep}{1mm}{
\begin{tabular}{c|cccccc}
\toprule
\textbf{public mean $f1$} & loss & batch size & accumulate steps& epochs & mixup & train+val \\ \midrule
0.76450    & Soft Target CE        & 28             & 2        &  150          & no           & no          \\
\bottomrule 
\end{tabular}
}
\end{table}

\begin{table}[htbp]
\caption{Mean $f1$ score on public test set with different accumulate steps.}
\label{table:2}
\setlength{\tabcolsep}{1mm}{
\begin{tabular}{c|cccccc}
\toprule
\textbf{public mean f1} & loss & batch size & accumulate steps & epochs & backbone & train+val \\ \midrule
0.76231    & Soft Target CE        & 28             & 1        &  150          & MetaFormer-2           & yes          \\
0.76450    & Soft Target CE        & 28             & 2        &  150          & MetaFormer-2           & yes          \\
0.76562    & Soft Target CE        & 28             & 4        &  150          & MetaFormer-2           & yes          \\
0.76601    & Soft Target CE        & 28             & 8        &  150          & MetaFormer-2           & yes          \\
\bottomrule 
\end{tabular}
}
\end{table}

\textbf{Network Framework. } MetaFormer compared with pure visual information input. In addition to MetaFormer, meanwhile we have tried to train ConvNext with image data. ConvNext’s results are listed in Table \ref{table:3}. The results of ConvNext are inferior to MetaFormer, thus we did not spend a lot of time to tuning it’s hyper-parameters, and we only add ConvNext-large to the model ensemble process.

\begin{table}[htbp]
\caption{Mean $f1$ score on public test set with different network framework.}
\label{table:3}
\setlength{\tabcolsep}{1mm}{
\begin{tabular}{c|cccccc}
\toprule
\textbf{public mean f1} & loss & batch size  & epochs & backbone & meta-information \\ \midrule
0.75238    & Soft Target CE        & 28        &  150          & MetaFormer-1           & no          \\
0.77056    & Soft Target CE        & 28        &  150          & MetaFormer-2           & no          \\
0.75563    & Soft Target CE        & 28        &  150          & ConvNext-base           & no          \\
0.76601    & Soft Target CE        & 28        &  150          & ConvNext-large          & no    \\  \midrule
0.76763    & Soft Target CE        & 28        &  150          & MetaFormer-1           &  yes          \\
0.78652    & Soft Target CE        & 28        &  150          & MetaFormer-2           & yes          \\
\bottomrule 
\end{tabular}
}
\end{table}

\textbf{Loss.} As shown in Table \ref{table:4}, Label Smoothing Cross Entropy without mixup augmentation
converges faster than Soft Target Cross Entropy with mixup augmentation, Seesaw Loss, Arcface Loss which designed for long tail recognition achieves better result on both public and private test set, we also attempt OHEM loss~\cite{ohem} try to possibly alleviate hard samples, a series of experiments show that it's not more effective.  

\begin{table}[htbp]
\caption{Mean $f1$ score on public test set with different loss functions.}
\label{table:4}
\setlength{\tabcolsep}{1mm}{
\begin{tabular}{c|cccccc}
\toprule
\textbf{public mean f1} & loss & batch size  & epochs & backbone & meta-information \\ \midrule
0.78652    & Soft Target CE        & 28        &  150          & MetaFormer-2           & yes          \\
0.83243    & Arcface                & 28        &  150          & MetaFormer-2           & yes          \\
0.82792    & Seesaw               & 28        &  150          & MetaFormer-2         & yes          \\
0.80569    & Polyloss        & 28        &  150          & MetaFormer-2         & yes   \\  
0.81478    & Circleloss        & 28        &  150          & MetaFormer-2         & yes   \\  
0.82733    & AM-softmax        & 28        &  150          & MetaFormer-2         & yes   \\  
0.83103    & OHEM        & 28        &  150          & MetaFormer-2         & yes   \\  
\bottomrule 
\end{tabular}
}
\end{table}

\textbf{Self-Supervised Learning. }We use MoCo method-base and Simclr method-base to compare, as shown in Table \ref{table:5}, some experiments show that Simclr method-based better than MoCo method-base, we analyzed that self-supervised representation learning joint with supervised learning is more friendly for downstream tasks,due to the semantic information learned by joint training has synchronized supervised information.  

\begin{table}[htbp]
\caption{Mean $f1$ score on public test set with self-supervised learning.}
\label{table:5}
\setlength{\tabcolsep}{1mm}{
\begin{tabular}{c|cccccc}
\toprule
\textbf{public mean f1} & loss & image size  & epochs & backbone & self-supervised \\ \midrule
0.85371    & Arcface        & 384        &  300          & MetaFormer-2           & Moco-method-base   \\
0.85734    & Arcface        & 384        &  300          & MetaFormer-2           & Simclr-method-base \\\midrule
0.86347    & Arcface        & 512        &  300          & MetaFormer-2           &  Simclr-method-base          \\
0.86035    & Arcface        & 384crop    &  300          & MetaFormer-2            & Simclr-method-base          \\
0.87741    & Arcface        & 384crop+aug &  300          & MetaFormer-2            & Simclr-method-base \\
\bottomrule 
\end{tabular}
}
\end{table}

\textbf{Pseudo label.} We approximately selected top scores test samples by it’s logit score, and
taken predicted class as their pseudo label. We trained MetaFormer-2 with train+val+pesudo, the results are listed in Table \ref{table:6}. 

\begin{table}[htbp]
\caption{Mean $f1$ score on public test set with pseudo label number.}
\label{table:6}
\setlength{\tabcolsep}{1mm}{
\begin{tabular}{c|cccccc}
\toprule
\textbf{public mean $f1$} & loss & image size & epochs & backbone & self-sup & pseudo label\\ \midrule
0.87803    & Arcface        & 384             & 150        &  MetaFormer-2          & yes   & +10\%   \\
0.88322    & Arcface        & 384             & 150        &  MetaFormer-2          & yes   & +20\%   \\
0.88767    & Arcface        & 384             & 150        &  MetaFormer-2          & yes   & +30\%   \\
0.88560    & Arcface        & 384             & 150        &  MetaFormer-2          & yes   & +40\%   \\
0.88591    & Arcface        & 384             & 150        &  MetaFormer-2          & yes   & +50\%   \\ \midrule
0.86697    & Arcface        & 384             & 30        &  MetaFormer-2          & yes   & +30\%   \\
0.90531    & Arcface        & 384             & 50        &  MetaFormer-2          & yes   & +30\%   \\ 
0.90324    & Arcface        & 384             & 60        &  MetaFormer-2          & yes   & +30\%   \\ 
0.89457    & Arcface        & 384             & 100        &  MetaFormer-2          & yes   & +30\%   \\
\bottomrule 
\end{tabular}
}
\end{table}

\textbf{Post Process}. As shown in Table \ref{table:7}, we use test time augmentation (TTA) trick in inference, and use final self-supervised Metaformer-2 with 384 image size, totally supervised Metaformer-2 with 384 image size. Meanwhile, combined self-supervised Metaformer-2 with 512 image size and totally supervised Metaformer-2 with 512 image size, also including ConvNext. We finally output fusion models with their highest logit score. It also our final result. 

\begin{table}[htbp]
\caption{Mean $f1$ score on public test set with post process method, including ensamble differenct models.}
\label{table:7}
\setlength{\tabcolsep}{1mm}{
\begin{tabular}{c|cccccc}
\toprule
\textbf{public mean $f1$} & model & size & crop  & TTA & epochs & add pseudo label  \\ \midrule
0.91324    & MetaFormer-2          & 384            & yes       &  yes         & 50          & yes          \\
0.91872    & MetaFormer-2          & 512             & yes        &  yes      & 50           & yes         \\
0.88356    & ssl+ MetaFormer-2     & 384             & no        &  yes          & (300,150)     & no          \\
0.89789    & ssl+ MetaFormer-2        & 512             & no       &  yes          & (300,150)   & no          \\\midrule
0.92778    & Ensamble          \\
\bottomrule 
\end{tabular}
}
\end{table}

\subsection{Other attempts}

\textbf{Batch Size. }Table \ref{table:8} illustrated that larger batch size improves the performance. in detail, by increasing batch size from 28 to 56, we improved $f1$ score from 0.76450 to 0.77211 on public set. 

\begin{table}[htbp]
\caption{Mean $f1$ score on public test set with different batch size.}
\label{table:8}
\setlength{\tabcolsep}{1mm}{
\begin{tabular}{c|cccccc}
\toprule
\textbf{public mean $f1$} & loss & batch size & accumulate steps& epochs & mixup & train+val \\ \midrule
0.76450    & Soft Target CE        & 28             & 2        &  32          & no           & no          \\
0.77211    & Soft Target CE        & 56             & 2        &  32          & no           & no          \\
\bottomrule 
\end{tabular}
}
\end{table}

\textbf{Training epochs. }We found the training epochs is not the more the better, as shown in Table \ref{table:9}, for MetaFormer-2, it is consistent that proper epochs is essential for better result. Following these observations, we fine-tune self-supvised MetaFormer-2 with 300 epochs, and trained supervised MetaFormer-2 with 180 epochs, which saved the training time.            

\begin{table}[htbp]
\caption{Mean $f1$ score on public test set with different training epochs.}
\label{table:9}
\setlength{\tabcolsep}{1mm}{
\begin{tabular}{c|cccccc}
\toprule
\textbf{public mean $f1$} & loss & batch size & accumulate steps& epochs & mixup & train+val \\ \midrule
0.72670    & Soft Target CE        & 28             & 2        &  50          & no           & no          \\
0.75891    & Soft Target CE        & 28             & 2        &  100          & no           & no          \\
0.76450    & Soft Target CE        & 28             & 2        &  150          & no           & no          \\ 
0.76231    & Soft Target CE        & 28             & 2        &  200          & no           & no          \\ \midrule
0.72882    & Soft Target CE        & 56             & 2        &  50          & no           & no          \\
0.76237    & Soft Target CE        & 56             & 2        &  100          & no           & no          \\
0.76284    & Soft Target CE        & 56             & 2        &  150         & no           & no          \\
\bottomrule 
\end{tabular}
}
\end{table}

\textbf{Image Size. }It is acknowledged that train the model with larger image size improves it’s performance. As there is a trade-off between image size and training flops, we have only tried to train MetaFormer-2 with 512 size, enlarge image size improves mean $f1$ score on private test set. 

\textbf{Pretrain Dataset. }We transfer MetaFormer-2 pretrained on different dataset such as imagenet22k and inaturalist21, the results are shown in Table \ref{table:10}. We think various pretrain provide
diverse single model, which will enhance the robustness of ensemble model.

\begin{table}[htbp]
\caption{Mean $f1$ score on public test set with different pre-trained datasets.}
\label{table:10}
\setlength{\tabcolsep}{1mm}{
\begin{tabular}{c|cccccc}
\toprule
\textbf{public mean $f1$} & loss & batch size & accumulate steps& epochs & mixup & pretain dataset \\ \midrule
0.76450    & Soft Target CE        & 28             & 2        &  150          & no           & imagenet22k          \\
0.76292    & Soft Target CE        & 28             & 2        &  150          & no           & inaturalist21          \\
\bottomrule 
\end{tabular}
}
\end{table}

\section{Conclusion}
In this paper, we have introduced our solution for SnakeCLEF 2022 competition. To solve this
challenging fine-grained visual classification, long-tailed distribution problem, we tried many efforts, such as different network framework to achieve more stronger feature extraction, loss function to solve long-tailed distribution and mine hard samples improve the robustness and generalization of model, self-supervised learning method fully mines unlabeled data information, designed special preprocess and post process tricks to improve overall performance. With these endeavours we achieved 1st place among the participants. The experimental results show the progressive process for single model, and the effectiveness of preprocess and post process for tail categories, other attempts also get expected gains. Due to limit of time and resource, more fine-grained experiments require further refinement. Meanwhile, we believe that meta-information is essential for fine-grained recognition tasks in the future. And, multimodal framwork would provide a way to utilize various auxiliary information. 

\section{Future Work}
For future work, firstly, it is valuable to study the method that self-supervised learning to fine-grained visual classification, we know that more and more research institutes and artificial intelligent companies like Google, Meta devoted to self-supervised representation learning recently, the trend of artificial intelligence towards general artificial intelligence, solved the problem of enormous labeled data. Finally, and the problem of distinguishing between tail categories is worth exploring.

\begin{acknowledgments}
    The authors would like to thank the organizer to offer a pratice and study platform. We also thank Dr Wei Zhu, Dr Yanming Fang for their guidance.
\end{acknowledgments}

\bibliography{sample-ceur}

\end{document}